# An Open-Source Dataset and A Multi-Task Model for Malay Named Entity Recognition

Yingwen Fu[1], Nankai Lin[1], Zhihe Yang[1], and Shengyi Jiang[1,2*]

[1] School of Information Science and Technology, Guangdong University of Foreign Studies, Guangdong, China
[2] Guangzhou Key Laboratory of Multilingual Intelligent Processing, Guangdong University of Foreign Studies, Guangzhou, China
`jiangshengyi@163.com`

**Abstract.** Named entity recognition (NER) is a fundamental task of natural language processing (NLP). However, most state-of-the-art research is mainly oriented to high-resource languages such as English and has not been widely applied to low-resource languages. In Malay language, relevant NER resources are limited. In this work, we propose a dataset construction framework, which is based on labeled datasets of homologous languages and iterative optimization, to build a Malay NER dataset (MYNER) comprising 28,991 sentences (over 384 thousand tokens). Additionally, to better integrate boundary information for NER, we propose a multi-task (MT) model with a bidirectional revision (Bi-revision) mechanism for Malay NER task. Specifically, an auxiliary task, boundary detection, is introduced to improve NER training in both explicit and implicit ways. Furthermore, a gated ignoring mechanism is proposed to conduct conditional label transfer and alleviate error propagation by the auxiliary task. Experimental results demonstrate that our model achieves comparable results over baselines on MYNER. The dataset and the model in this paper would be publicly released as a benchmark dataset.

## 1 Introduction

As the official language of Malaysia, Brunei and Singapore, Malay is a member of the Malayo-Polynesian branch in the Austronesian language family. Faced with massive amounts of raw data in Malay, it is particularly significant to utilize natural language processing (NLP) technology to extract pattern information. With the recent rapid development of deep learning, many researchers have attempted to utilize neural approaches for various NLP tasks. However, neural approaches rely heavily on abundant labeled data, thus they are mainly oriented to high-resource languages such as English. Malay is a low-resource language for which labeled data resource is scarce and NLP technology is not yet well-developed. Most existing studies are still rule-based and machine learning-based (ML) due to data-hungry limitation.

[*] Corresponding author. E-mail: jiangshengyi@163.com.

Named Entity Recognition (NER) is fundamental to NLP. It is designed to identify the boundaries of entities and classify them into predefined categories. Various neural approaches for NER have been proposed in recent years. One representative direction is to incorporate more useful features into an NER system, such as part-of-speech (POS) labels and boundary information [2-8]. We could see potential benefits of these external features for NER models. However, how to better integrate these features to further improve the NER performance calls for better strategies, models and the like. Besides, boundary recognition error (BRE) is also one of the important factors affecting the performance of NER [8]. Therefore, a solution to the BRE problem is urgently needed.

It is natural that some low-resource languages have little labeled resource, while their homologous languages have certain labeled resources, and there are great similarities between homologous languages such as Spanish-Portuguese and Indonesian-Malay [26]. Therefore, a potential dataset construction method is borrowing from the datasets of homologous languages. To alleviate the current insufficiency of language resources in Malay, we propose a dataset construction framework to build a Malay NER dataset (MYNER) comprising 28,991 sentences. Specifically, the framework consists of two parts: (1) **preliminary dataset construction** based on labeled datasets of homologous languages (Indonesian) and rules and (2) **iterative optimization** that is a semi-manual method to iteratively optimize the dataset based on NER models and manual audition. **It is worthwhile to note that the proposed framework works for the languages whose homologous languages have labeled datasets.** In addition, to tackle the BRE problem, we propose a neural multi-task (MT) model with a bidirectional revision (Bi-revision) mechanism (MTBR). Instead of auxiliary features, we treat boundary detection (BD) as an auxiliary task and make use of the mutual relations in two tasks in a more advanced and intelligent way: (1) our method benefits from general representations of both tasks provided by multi-task learning, which enjoys a regularization effect [9-10] that can well minimize over-fitting to NER. (2) we leverage explicit boundary information to guide the model to locate and classify entities precisely. The Bi-revision mechanism in MTBR is constructed that the label probabilities of the BD task would explicitly revise the label probabilities of the NER task (main task). (3) A gated ignoring mechanism is proposed in MTBR to alleviate error propagation by the BD task. It transforms from the hidden features of NER module and determines the degree of the auxiliary task revision. Furthermore, a random probability is introduced to control whether the gate mechanism would take effect. Two tasks are trained in an alternate manner.

The main contributions of this paper are presented as follows:

1) A dataset construction framework which based on labeled datasets of homologous languages and iterative optimization is proposed to construct Malay NER dataset. The framework can be applied to other homologous languages.

2) A large-scale and high-quality Malay NER dataset (MYNER) is constructed and would be publicly available as a benchmark dataset.

3) A neural multi-task (MT) framework with a bidirectional revision (Bi-revision) mechanism is presented to tackle the Malay NER task (MTBR).

4) Based on MYNER, MTBR achieves competitive performance compared with multiple single-task and multi-task baselines.

## 2      Related Work

**NER.** In recent years, there has been an increasing amount of research on NER. Traditional approaches to NER include handcrafted features represented by Hidden Markov Models [27] and Conditional Random Fields (CRF) [2]. State-of-the-art neural NER techniques use a combination of pre-trained word embeddings [11-12] and character embeddings derived from a convolutional neural network (CNN) layer or bidirectional long short-term memory (Bi-LSTM) layer [3-8]. These features are passed to a Bi-LSTM layer, which may be followed by a CRF layer [13-15]. Recently, fine-tuned pre-trained language models (PLMs) like ELMO [17], Flair [28] and BERT [16] are also popular for NER. But they are mainly oriented to label-rich languages such as English.
**Malay NER.** Most current research on Malay NER is based on rules and ML. Alfred [1] used a rule-based approach to identify named entities in Malay texts. This work centered on designing rules based on the POS tagging features and contextual features, which reached the F-Measure value of 89.47%. Ulanganathan [18] used CRF to construct a Malay language named entity recognition engine (Mi-NER) based on a POS tagger. The F-Measure of this work was 74.98%. Furthermore, Sulaiman [19] conducted an experiment to detect Malay named entities. Stanford NER and Illinois NER tools were used in this work to identify Malay named entities. Asmai [20] presented an enhanced Malay named entity recognition model using combination of fuzzy c-means and K-nearest neighbors algorithm method for crime analysis. **They are all based on different non-public datasets and there is no authoritative benchmark for the Malay NER task.**
**Multi-Task Learning.** As a research hotspot in recent years, multi-task learning has been applied successfully across many applications of artificial intelligence (AI), from NLP and speech processing to computer vision [10]. It enables models to generalize better on the original task by sharing representations between related tasks. One of the multi-task learning methods for deep learning is hard parameter sharing of hidden layers. It is generally applied by sharing the hidden layers between all tasks, while retaining several task-specific output layers [21].

## 3      Dataset Construction

High-quality datasets are the foundation of the NLP technology development. However, as far as we know, there are currently no open-source Malay NER datasets. In addition, fully human annotated datasets for NER are expensive and time-consuming which are therefore relatively small. To reduce the burden of manual annotation, a semi-manual method is proposed to construct a Malay NER dataset which would be publicly available with the publication of this paper.

### 3.1      Dataset Design

Indonesian and Malay are close languages that have high lexical similarity [26]. Based on this language feature, we propose to construct a Malay preliminary dataset from the existing Indonesian datasets [22-23]. In addition, some rules [1] are leveraged to detect entities in Malay texts to expand the preliminary dataset. After that, a semi-manual method is proposed to iteratively update the dataset and the model (introduced in Section 4). Totally, the dataset contains 28,991 sentences (384010 tokens) with three entity categories (**person, location and organization**). Some statistics are shown in Table 1. The dataset is divided into a training set (80%), a validation set (10%) and a testing set (10%) for subsequent experiments in BIO2 format.

**Table 1.** Statistics of MYNER

| Entity Type | Abbreviation | Number |
|---|---|---|
| Person | PER | 37,473 |
| Location | LOC | 20,234 |
| Organization | ORG | 19,646 |

### 3.2 Data Source

We use Malay news articles to construct the dataset because of its convenient availability and huge amount. We crawl articles from Malay news websites whose content covers various topics including politics, finance, society, military, etc. The websites are shown in Table 2. After separating paragraphs into individual sentences, we randomly select 30,000 sentences to be annotated. Besides, we construct a large Malay vocabulary with 81,691 tokens based on the news articles.

**Table 2.** Data Source

| Website | URL |
|---|---|
| Bharian | http://www.bharian.com.my/ |
| Utusan | http://www.utusan.com.my/ |
| Malaysiakini | http://www.malaysiakini.com/bm |

### 3.3 Dataset Construction

The construction process consists of two parts: preliminary construction and iterative optimization.

(1) Collect the Indonesian NER datasets from the existing works [22-23] and separate them into individual sentences. Search whether all tokens of each sentence exist in the Malay dictionary. If so, add the sentence to the preliminary dataset, otherwise, discard it.
(2) According to the rules defined in [1], expand the preliminary dataset within the unlabeled Malay sentences.

Through the above two steps, a seed dataset is constructed consisting of 28,991 sentences. However, the dataset constructed in these ways often suffers from mislabeling errors. Therefore, we propose the following steps to optimize the dataset.

(3) Use all annotated sentences (28,911 sentences) for both model training and testing.
(4) Manually audit the sentences with different labels in training and testing phases. For each sentence to be audited, two auditors are engaged in to ensure the quality of the audited sentences. If two audit results conflict, ask another auditor to check the results.

5(5) Re-train the model with the audited dataset.
(6) Repeat step (3) – (5) until the dataset and the model convergence.

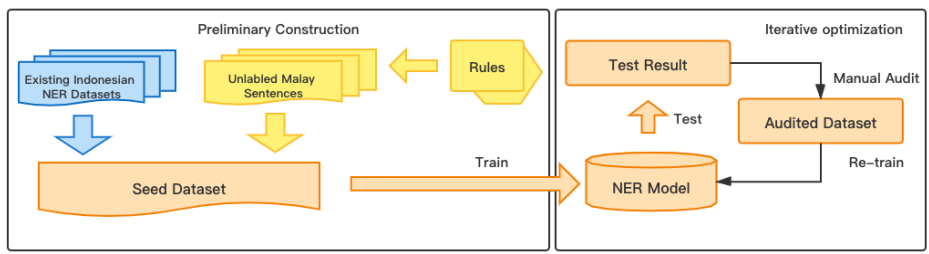

**Fig. 1.** Dataset construction process which consists of two parts: preliminary construction and iterative optimization.

## 4 Model

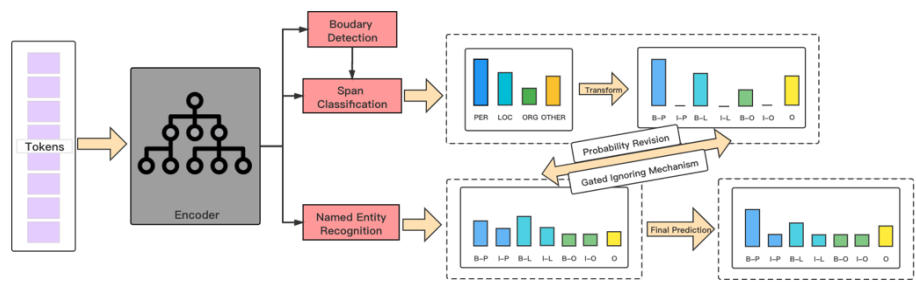

**Fig. 2.** The MTBR model structure. Due to space limitation, [B-P, I-P, B-L, I-L, B-O, I-O, O] in the figure represents [B-PER, I-PER, B-LOC, I-LOC, B-ORG, I-ORG, OTHER].

### 4.1 Overview

Consider a sentence $S$ with $L$ tokens $\{w_i\}_{i=1}^{L}$, we respectively assign each token $w_i$ with start label $y_s^i$ and end label $y_e^i$ for BD task (token-wise binary classification task) and entity label $y_{ner}^i$ for NER task. In addition, we introduce a span classification for BD module aiming to assign one of four entity tags (**PER, LOC, ORG** and **O**) $y_{au\_ner}^i$ for span representations.

Following the overview in Figure 2, MTBR in this paper is a multi-task model consisting of BD task and NER task. BD module is meant to predict the beginning or the end of an entity, along with a task that classifies the spans as corresponding entity tags. The token representation from the encoder is universally shared in the downstream two tasks to minimize over-fitting to each task. In addition to implicitly improving the contextual representation of NER through the multi-task model, the span classifier is designed to explicitly revise the NER outputs through a gated ignoring mechanism. Two tasks are trained in an alternate manner.

## 4.2 Encoder

We utilize the pre-trained BERT [16] as our encoder for contextual representations $h_i$. It is noted that other network structures such as LSTM are also suitable for the encoder here. It is formulated as:

$$h_i = BERT(w_i) \tag{1}$$

After obtaining the token representations from the encoder, we apply two separate MLPs to create different representations ($h_i^{bd}$/$h_i^{ner}$) for the (BD/NER) modules. Using different representations allow the system to learn from two tasks separately.

$$h_i^{bd} = MLP_{bd}(h_i); \quad h_i^{ner} = MLP_{ner}(h_i) \tag{2}$$

## 4.3 Boundary Detection Module

**Boundary Detection.** Instead of treating boundary detection as a sequence labeling task, we predict the start and end positions with two token-wise classifiers. The contextual representation $h_i^{bd}$ is fed into a MLP classifier to predict the boundary labels of $w_i$.

$$p_s^i = softmax(W_s \cdot h_i^{bd} + b_s); \quad p_e^i = softmax(W_e \cdot h_i^{bd} + b_e) \tag{3}$$

Where $W_s$ and $W_e$ are fully connected matrices, $b_s$ and $b_e$ are bias vectors.

**Span Classification.** After obtaining the boundaries of entities, in order to better prompt the training of the NER task, we further introduce a task that classifies the spans as corresponding entity tags. We define consecutive tokens between the nearest pair of the start and end boundaries as an entity to be labeled, and the tokens outside the boundaries as non-entities. We calculate the summarized span representation $v_{sp}$ and the label $y_{sp}$ by averaging the task-specific representations and labels in their corresponding boundaries $(i, j)$.

$$v_{sp} = \frac{1}{j-i+1}\sum_{t=1}^{j} h_t^{bd}; \quad y_{sp} = \frac{1}{j-i+1}\sum_{t=1}^{j} y_{au\_ner}^t \tag{4}$$

Later, the entity representation $v_{sp}$ is fed into a MLP classifier to predict its entity tag.

$$p_{sp} = softmax(W_{sp} \cdot v_{sp} + b_{sp}) \tag{5}$$

**Loss.** For boundary detection, we minimize two cross-entropy (CE) losses of the beginning and end boundaries $\mathcal{L}_{bd} = \mathcal{L}_{bd}^s + \mathcal{L}_{bd}^e$ and a CE loss $\mathcal{L}_{sp}$ for span classification. The total loss is the weighted sum of two losses, and $w_1$ in this paper is 0.5.

$$\mathcal{L}_1 = w_1 \mathcal{L}_{bd} + (1 - w_1)\mathcal{L}_{sp} \tag{6}$$

## 4.4 NER Module

**Standard NER.** As a common task of sequence labeling, the contextual representation $h_i^{ner}$ for NER module is fed into a MLP classifier to predict the entity tag of $w_i$.

$$p_{ner}^i = softmax(W_{ner} \cdot h_i^{ner} + b_{ner}) \quad (7)$$

Where $W_{ner}$ is a fully connected matrix, $b_{ner}$ is a bias vector.

**Bi-revision Mechanism.** In addition to implicitly improving the NER performance, we assume that the output probabilities of span classification can further revise the result of the NER module. Conversely, to alleviate the error propagation caused by the BD task, the NER module is conducive to verifying the revision of the BD task. Thus, we utilize the label probabilities of the span classification through a gated ignoring mechanism to obtain an adjusted probability. Specifically, we first need to transform the label probability of the subtask as in Figure 3. For each token, the span classification module outputs the probability value $p_{sp} = [p_{PER}, p_{LOC}, p_{ORG}, p_O]$ of $y_{au\_ner}^i$. Each type of $y_{au\_ner}^i$ except $O$ (other) label corresponds to two $y_{ner}^i$ labels (for instance, PER label corresponds to B-PER label and I-PER label). So, we transform $p_{sp}$ to $p_{sp\_new} = [p_{B-PER}, p_{I-PER}, p_{B-LOC}, p_{I-LOC}, p_{B-ORG}, p_{I-ORG}, p_O]$. If $w_i$ is the first token of the detected entity, the values of $(p_{PER}, p_{LOC}, p_{ORG}, p_O)$ are assigned to $(p_{B-PER}, p_{B-LOC}, p_{B-ORG}, p_O)$, or they are assigned to $(p_{I-PER}, p_{I-LOC}, p_{I-ORG}, p_O)$. And the remaining tag probabilities are set to 0. It is noted that we set all label probabilities to 0 for those non-entities, so they will not affect the label probability distribution of the NER module. After obtaining the transformed probability $p_{sp'}^i$, we calculate the revised NER probability as:

$$p_{ner'}^i = p_{ner}^i + gate_i \cdot p_{new\_sp}^i \quad (8)$$

**Gated Ignoring Mechanism.** The gated ignoring mechanism is constructed to alleviate the error propagation caused by the span classification task. It consists of two components: a gate mechanism that determines the degree of probability revision:

$$gate_i = sigmoid(W_g \cdot h_i^{ner} + b_g) \quad (9)$$

where $W_g$ is a fully connected matrix, $b_g$ is a bias vector,
and a random probability $p$ (with a threshold $\alpha$) to control the gate mechanism:

$$p_{final\_ner}^i = \begin{cases} p_{ner}^i, & p > \alpha \\ p_{ner'}^i, & otherwise \end{cases} \quad (10)$$

**Loss.** We minimize a cross-entropy loss $\mathcal{L}_3$ on $p_{final\_ner}^i$ for NER task.

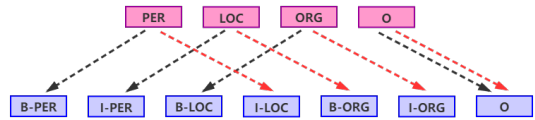

**Fig. 3.** Probability Transformation. If $w_i$ is the first token of the detected entity, the probabilities would flow towards the black arrow, otherwise they would flow towards the red arrow.

## 5 Experiment

### 5.1 Baseline Models

Most current Malay NER models are based on rules and ML. Because of the high cost and poor portability of them, Malay NER methods would not be used as baselines in this paper.

We compare MTBR with some state-of-the-art neural NER models (CNNs-Bi-LSTM-CRF [15], BERT [16], ELMO [17], and some multi-task models[1] (Bi-LSTM-PN [6], MT-BERT [25], MT-MED [24]). We pre-train the ELMO model with massive Malay news articles and leverage Bahasa-BERT[2] as the encoder of the MT-BERT baseline model and the proposed model. These baselines are evaluated on MYNER. We repeat each experiment 5 times and report the average performance on the test set. The evaluation metrics are accuracy, recall and F1-score.

### 5.2 Model Settings

Our model is based on PyTorch[3] (BERT models are based on HuggingFace[4]). The model settings are shown in Table 3.

**Table 3.** Model Settings

| Model | Component | Parameter | Value |
|---|---|---|---|
| CNNs-Bi-LSTM-CRF | CNN | Filter Size | 3 |
| | | Number of Filters | 30 |
| | Bi-LSTM | Number of Units | 200 |
| | Optimizer | Optimizer Function | SGD |
| | | Learning Rate | 0.01 |
| | Word Embedding | Word Embedding | GloVe [12] |
| | | Dimension | 300 |
| ELMO | Optimizer | Optimizer Function | SGD |
| | | Learning Rate | 0.015 |
| | | Learning Rate Decay | 0.05 |
| BERT | Optimizer | Optimizer Function | Adam |
| | | Learning Rate | 5e-5 |
| | | Weight Decay | 0.001 |
| - | Common | Max Length | 128 |
| | | Data Source | Malay News Articles |

---

[1] It is worthwhile to note that we follow the structure of these works and train BD and NER tasks jointly.
[2] https://huggingface.co/huseinzol05/bert-base-bahasa-cased
[3] https://pytorch.org/
[4] https://huggingface.co/transformers/

## 5.3 Main Performance

**Table 4.** Main Performance between other methods and MTBR.

| Framework | Method | P (%) | R (%) | F1 (%) |
|---|---|---|---|---|
| Single-task Model | Bi-LSTM-CNNs-CRF [15] | 88.73 | 88.12 | 88.42 |
| | ELMO | 91.12 | 89.67 | 90.39 |
| | mBERT[5] [16] | 91.02 | 89.55 | 90.28 |
| | Bahasa-BERT[6] [16] | 91.01 | 90.11 | 90.56 |
| Multi-task Model | Bi-LSTM-PN[7] [6] | 88.25 | 88.46 | 88.36 |
| | MT-MED [24] | 89.57 | 88.85 | 89.20 |
| | MT-BERT [25] | 91.80 | 90.24 | 91.01 |
| | MTBR (Bi-LSTM-CNNs-CRF) | 89.91 | 89.69 | 89.80 |
| | MTBR (ELMO) | **92.97** | 91.44 | 92.20 |
| | MTBR (Bahasa-BERT) | 92.70 | **92.01** | **92.36** |

The first part of Table 4 illustrates the results of 4 previous top-performance single-task systems for NER. Among these previous studies, fine-tuned language models (ELMO, mBERT and Bahasa-BERT) significantly outperforms classic neural models represented by Bi-LSTM-CNNs-CRF. In addition, it is worthwhile to note that ELMO works slightly better than mBERT, we hold the idea that ELMO is Malay-oriented that is pre-trained in Malay news articles whereas mBERT is pre-trained in corpora of multiple languages and focuses on learning language-independent knowledge, leading to insufficient language-specific knowledge for Malay NER. Besides, Bahasa-BERT achieves best performance in four single-task models, so we leverage it as our encoder for the follow-up experiments.

The second part of Table 2 shows the effectiveness of some multi-task models. Among them, MT-MED and Bi-LSTM-PN leverage Bi-LSTM-CNNs as encoder and MT-BERT is based on BERT. We can see that MT-BERT outperforms Bi-LSTM-PN and MT-MED which should be credited to pre-trained language models represented by BERT.

MTBR obtains the best performance by improving more than 1.3% over the best baseline model MT-BERT. Because of the novel design of Bi-revision mechanism, MTBR can utilize not only the general representations of different tasks, but also label probabilities of different tasks. Hence, MTBR obtains significant improvements compared with multi-task models and single-task models. Meanwhile, we could see that the MTBR with Bahasa-BERT model as encoder achieves state-of-the-art performance, which could be regarded as the new baseline method.

---

[5] https://huggingface.co/bert-base-multilingual-cased
[6] https://huggingface.co/huseinzol05/bert-base-bahasa-cased
[7] We reproduce this work and add one softmax layer for the NER task.

### 5.4 Ablation Study

**Analysis of Different Components.** To evaluate the effectiveness of different components in MTBR, we remove a certain part of the model for experimentation. As Table 5 shows, within the multi-task framework, BD task contributes to boost the NER performance with an improvement of 1.8% in F1-score, which should be credited to its BRE correction capability. Besides, Bi-revision module and the gated ignoring mechanism make great contributions to MTBR model because the Bi-revision mechanism can effectively revise the NER output probability, and the gated ignoring mechanism can verify the correctness of the revision and control the degree of revision of the auxiliary task. It can well alleviate error propagation. Moreover, the random probability in the gated ignoring mechanism can further reduce error propagation. It seems that MTBR can make full use of the advantages of multi-task learning from different perspectives.

Table 5. Performance of Different Modules

| Model | P (%) | R (%) | F1 (%) |
|---|---|---|---|
| w/o Boundary Detection | 91.01 | 90.11 | 90.56 |
| w/o Bi-Revision | 91.80 | 90.24 | 91.01 |
| w/o Gated Ignoring Mechanism | 92.70 | 90.89 | 91.79 |
| w/o Random Probability | 92.66 | 91.34 | 92.00 |
| MTBR | **92.70** | **92.01** | **92.36** |

**Boundary Recognition Error Analysis.** Since we explicitly introduce an auxiliary task of boundary recognition, BRE tends to decline. Table 6 compares the ratios of BRE in different models. It is clear that our proposed MTBR model with BD auxiliary task can significantly reduce BRE, thus improving the NER performance.

Table 6. Boundary Recognition Error Analysis

| Model | BRE (%) |
|---|---|
| Sigle-task Model | 3.53 |
| Standard Multi-task Model | 2.79 |
| MTBR Model | **1.89** |

## 6 Conclusion

In this paper, we propose a dataset construction framework based on labeled datasets of homologous languages and iterative optimization to construct a Malay named entity recognition (NER) dataset (MYNER) using a large amount of news articles for Malay NER task. The proposed framework works for the languages whose homologous languages have labeled datasets. In addition, previous studies demonstrate the potential benefits of boundary detection (BD) task for NER. Based on MYNER, we further explore how the BD task can better improve NER performance and propose a neural multi-task model with a bidirectional revision (Bi-revision) mechanism (MTBR) taking both model transfer and label transfer of BD task into account with some probabilities to improve the NER performance. Experimental results show that MTBR can obtain competitive performance over multiple baselines in MYNER. The dataset and the model would be publicly available as a new benchmark.